\documentclass[11pt,a4paper]{article}
\usepackage{aaai19}
\usepackage{times}
\usepackage{latexsym}

\usepackage{url}

\usepackage{CJKutf8}
\usepackage{amsmath}
\usepackage{graphicx}
\usepackage{multirow}
\usepackage{amssymb}
\usepackage{booktabs}
\usepackage{todonotes}
\usepackage{verbatim}
\usepackage{caption}
\usepackage{subcaption}
\usepackage{amsmath}
\usepackage{algorithm}
\usepackage{algpseudocode}

\DeclareMathOperator*{\argmax}{argmax}

\newcommand{\newcite}[1]{\citeauthor{#1} (\citeyear{#1})}

\title{Chat More If You Like: Dynamic Cue Words \\Planning to Flow Longer Conversations}

\author{Lili Yao$^{1}$\thanks{Equal contribution: Lili Yao and Ruijian Xu}, Ruijian Xu$^{1*}$, Chao Li$^2$, Dongyan Zhao$^1$ and Rui Yan$^1$ \\
\texttt{\{yaolili,xurj,zhaodongyan,ruiyan\}@pku.edu.cn}\\
\texttt{lichao17@xiaomi.com}\\
$^1$ Institute of Computer Science and Technology, Peking University\\
$^2$ Beijing Xiaomi Mobile Software
}

\begin{document}
\begin{CJK*}{UTF8}{gkai}

\maketitle
\begin{abstract}
To build an open-domain multi-turn conversation system is one of the most interesting and challenging tasks in Artificial Intelligence. Many research efforts have been dedicated to building such dialogue systems, yet few shed light on modeling the conversation flow in an ongoing dialogue. Besides, it is common for people to talk about highly relevant aspects during a conversation. And the topics are coherent and drift naturally, which demonstrates the necessity of dialogue flow modeling. To this end, we present the multi-turn cue-words driven conversation system with reinforcement learning method (RLCw), which strives to select an adaptive cue word with the greatest future credit, and therefore improve the quality of generated responses. We introduce a new reward to measure the quality of cue words in terms of effectiveness and relevance. To further optimize the model for long-term conversations, a reinforcement approach is adopted in this paper. Experiments on real-life dataset demonstrate that our model consistently outperforms a set of competitive baselines in terms of simulated turns, diversity and human evaluation.


\end{abstract}

\section{Introduction}
\label{intro}
Building a conversational system that enables natural human-computer interaction has been more and more important. Previous efforts focus on task-oriented dialogue systems \cite{wen2016network,eric2017copy,liu2017end} which help people complete specific tasks in vertical domains. Recently, non-task-oriented dialogue systems \cite{higuchi2008casual,yu2016strategy} that converse with humans on open domain topics are attracting increasing attention, due to their various applications, such as chatbots, personal assistants, and interactive question answering etc.

Basically, there are two major categories of open-domain conversation systems: single- and multi-turn dialogue systems. For single-turn dialogue systems, previous research~\cite{shang2015neural,vinyals2015neural,dai2015semi,li2016diversity,li2016persona,mou2016sequence,xing2016topic,vougiouklis2016neural,yao2017towards} concentrated on generating a relevant and diverse response when given static context. One of the significant issues is that these systems often generate universal responses such as ``I don't know'' and ``Okay''~\cite{li2016diversity,serban2016building,mou2016sequence}. Besides, the single-turn dialogue systems ignore the long-term dependency among generated responses that is critical in natural conversation. To build a natural and coherent conversation interface, the multi-turn dialogue systems are currently the primary choice. To enhance the long-term dependency modeled in multi-turn dialogue systems, reinforcement learning based dialogue generation methods~\cite{li2016deep,asghar2016online,dhingra2016end} are proposed. Nevertheless, the performance of existing conversation systems is still far from satisfactory. 

\begin{table}[t!]
  \resizebox{0.92\columnwidth}{!}{
  \begin{tabular}{p{2.4cm}|p{5cm}}
  \toprule
	Cue word & Utterance  \\
  \midrule
-  & A: 去哪里(Where are you going?) \\
回家(home)  & B: 回家(Home.) \\
上班(working)  & A: 好吧，我还在上班(Well, I am still working.)\\
加班(overtime)  & B: 加班？你太辛苦了(Work overtime? You are too hard.)\\
委屈(aggrieved)  & A: 我也很委屈(I feel aggrieved.)\\
  \bottomrule
  \end{tabular}}
  \caption{An example of cue words and utterances in a conversation.}
  \label{cuewords}
\end{table}

In human-human conversations, people tend to talk about highly relevant aspects and topics during a chat session. To make dialogues more interesting, they will find satisfying topics dynamically. It is easy for them to recognize key signs of discomfort, which can be a juncture to seek a new topic. Also, such implicit information~\cite{yao2017towards} has proven effective for meaningful responses generation.

However, it is difficult and challenging to launch such a human-computer conversation system.  
1) Although topics augmented neural response generation methods~\cite{mou2016sequence,yao2017towards,wang2018chat} have shown impressive potential in single-turn conversation system, they do not apply to ongoing dialogues. Because they ignore the long-term impact of the selected cue words. Besides, the selection of cue words is based on a specific measurement, such as Point-wise Mutual Information (PMI)~\cite{mou2016sequence,yao2017towards}, or direct extraction of important words from context~\cite{wang2018chat}, which is not trainable in ongoing dialogues. 
2) It is complicated to model the practical flow of a real conversation. Usually, the conversational topics are coherent and drift naturally. However, if the topic makes both sides of the conversation feel uncomfortable, they may try to change the topic.

To tackle these issues, we present a multi-turn cue words driven conversation system with reinforcement learning, named \textit{RLCw}. Specifically, we aim to model the topic flow of an ongoing dialogue with cue words. In each turn, we strive to select an adaptive cue word with the greatest future credit based on the dialogue state (history context and cue words). Further, we take the cue words as the main gist of the upcoming utterances to guide the response generation. As shown in Table~\ref{cuewords}, the selected cue words dynamically drive the dialogue direction and help to generate an informative and interesting conversation. Main contributions of this paper include: 


\begin{itemize}
\item In a multi-turn dialogue, we adopt cue words to shape the conversation flow, and unify cue words prediction and responses generation in an end-to-end framework.
\item We propose to measure the quality of a cue word from two aspects: effectiveness and relevance. In this way, the selected cue words with higher reward could further drive the dialogue to be more informative and flow a longer and more fluent conversation.
\item Extensive comparisons and analyses are conducted to draw insights into how our proposed RLCw model lead the conversation to a better direction.
\end{itemize}

The rest of this paper is organized as follows. We firstly review previous related work. Next, we present the overall framework and describe the proposed methods in detail. This is followed by model training process. Finally, we elaborate our experimental setup, results, analysis, and draw our conclusion.
\section{Related Work}  
\label{relatedwork}

To build an open-domain conversation system is one of the most interesting and challenging topics in both artificial intelligence and natural language processing research these years. 
For single-turn conversation systems, prior studies strived to generate more meaningful and informative responses. There are three mainstream ways to address this issue. 1) To modify loss function or to improve the beam search algorithm. \newcite{li2016diversity} proposed to use Maximum Mutual Information (MMI) as the objective function in neural models. \newcite{shao2017generating} introduced a stochastic beam-search algorithm with segment-by-segment re-ranking and injected diversity in generation process earlier. 2) To learn latent variables. \newcite{zhao2017learning} adopted an utterance-level latent variable to model the distribution of the next response so that the system could generate more diverse responses. 3) To fuse additional information. \newcite{mou2016sequence} leveraged the Pointwise Mutual Information (PMI) to predict a keyword and presented the seq2BF framework to generate a reply containing the given keywords. \newcite{yao2017towards} proposed an implicit content-introducing method to incorporate keyword information in a soft schema. Besides, topic information, which is regarded as prior knowledge, has been shown effective in conversation systems~\cite{xing2016topic}. Inspired by these studies, we also resort to enlightening cue words to improve the informativeness and meaningfulness of generated responses.

As for multi-turn conversation systems, \newcite{serban2016building} presented a hierarchical recurrent encoder-decoder (HRED) approach to encode each utterance and to recurrently model the dialogue context to generate responses, which was further improved through a stochastic latent variable at each dialogue turn~\cite{serban2017hierarchical}. These works focused on the static dialogue context.
In an ongoing dialogue, naturally, deep reinforcement learning method has been used to improve the performance of response generation. \newcite{li2016dialogue} explored an online learning fashion: the system learned from the feedback of the dialogue partner. \newcite{asghar2016online} proposed an active learning approach to learn user explicit feedback online and to combine the offline supervised learning for response generation of conversational agents. ~\newcite{dhingra2016end} presented an end-to-end dialogue system for information acquisition from a knowledge base using reinforcement learning. \newcite{li2016deep} attempted to model the future influence of generated responses using a deep reinforcement learning approach to optimize generation model. Based on this, \newcite{zhang2018exploring} also added the implicit feedback as a part of the reward. 

Different from existing works, our goal is to model the future direction of ongoing conversations. To achieve this, we designed our model to dynamically selected cue words and integrate them into the decoder to generate proper replies in multi-turn conversations, thus improving our model in terms of user engagement.

\section{Methods}
\label{method}



\begin{figure}[t]
  \centering
  \resizebox{0.8\columnwidth}{!}{
  \includegraphics[width=.40\textwidth]{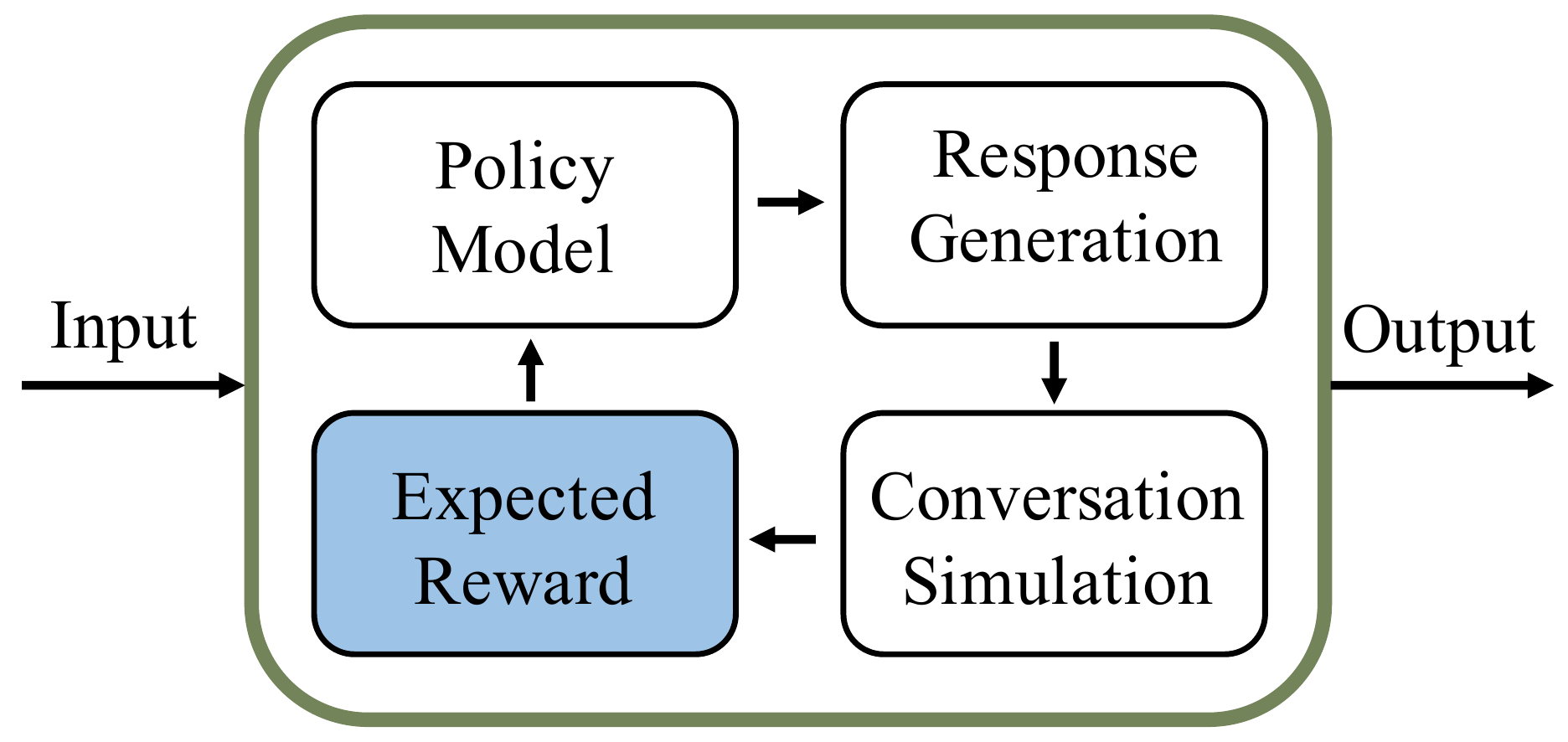}
  }
  \caption{The pipeline of our proposed RLCw system. }
  \label{overview}
\end{figure}

\subsection{System Overview}
In this paper, our goal is to dynamically shift conversations to better topics and make conversations more attractive. 
Figure~\ref{overview} illustrates the pipeline of our proposed RLCw system. We treat the cue word prediction as an action that is taken according to a policy. Given the source input, the policy model firstly samples a relevant cue word. Then a response will be generated based on both dialogue state and the selected cue word. After that, we estimate the expected reward in the following conversation and optimize the parameters of the policy network.

Formally, a dialogue session $\mathcal{S}_i=(u_{i,1}, u_{i,2}, ..., u_{i,j-1})$ and history cue words sequence $\mathcal{F}_i= (a_{i,1}, a_{i,2}, ..., a_{i,j-1})$ are given to the system, where $u_{i,k}$ is the $k$-th turn's utterance in the $i$-th session, and $a_{i,k}$ denotes for the cue word corresponding to $u_{i,k}$. Based on the dialogue state, the system firstly selects an adaptive cue word $a_{i,j}$ using policy model, and then generates a reply $R$ with cue word augmented response generation framework. The details are as follow.

\subsection{Cue Word Augmented Response Generation} 
We employ neural sequence to sequence model for our response generation. In this paper, we adopt the two LSTM~\cite{hochreiter1997long} layers framework~\cite{venugopalan2015sequence}, which shares parameters between encoder and decoder module. 

To generate a response $R=(y_1,...,y_m)$, we maximize the generation probability conditioned on input query $Q=(x_1,...,x_n)$ and selected cue word $a_{i,j}$. Due
to the computational complexity, the input query is the concatenation of previous two utterances $Q = [u_{i,j-2}, u_{i,j-1}]$.

At time $t$, the encoding hidden state of the first layer $l_1$ and second layer $l_2$ are defined as:
\begin{equation}
\label{encoder}
\begin{aligned}
h_{t}^{l_1} &= \textsc{LSTM}\left(x_{t}, h_{{t}-1}^{l_1}\right) \\
h_{t}^{l_2} &= \textsc{LSTM}\left(\left \langle pad \right \rangle, h_{t}^{l_1}, h_{{t}-1}^{l_2}\right)
\end{aligned}
\end{equation}
where $x_t$ is the input word embedding. 
The special symbol $\left \langle pad \right \rangle$ denotes padding with zero. $h_0^{l_1}$ and $h_0^{l_2}$ are initialized with zero vectors.

 During decoding process, we initialize the decoder with final states of encoder. And then, the decoder generates reply words one by one.
 To incorporate the predicted cue words into generation process, we introduce cue word information at every step in decoding inspired by~\newcite{yao2017towards}. At generation time step $t$, the decoder hidden state of two layers are given by:



\begin{equation}
\label{decoder}
\begin{aligned}
s_{t}^{l_1} &= \textsc{LSTM}\left(I, s_{{t}-1}^{l_1} \right) \\
s_{t}^{l_2} &= \textsc{LSTM}\left(y_{{t}-1}, s_{t}^{l_1}, s_{{t}-1}^{l_2} \right) \\
p_{t} &= \textit{softmax}\left(\eta\left(s_{t}^{l_2}\right)\right)
\end{aligned}
\end{equation}
where $y_{t-1}$ is the output word embedding last time, $p_{t}$ denotes the probability distribution of candidate words at generation time step $t$. $\eta$ is implemented as a multi-layer perceptron (MLP) layer. $I$ denotes cue words information, which is the linear transformation of cue word embedding:

\begin{equation}
\label{cue word fusion}
\begin{aligned}
I = W_I a_{i,j} + b_I
\end{aligned}
\end{equation}
where $W_I$ and $b_I$ are weight matrices and bias terms, respectively. In this way, we fuse cue words information into generation process so that the system is aware of dialogue direction.

\begin{figure}[t]
  \centering
  \includegraphics[width=.47\textwidth]{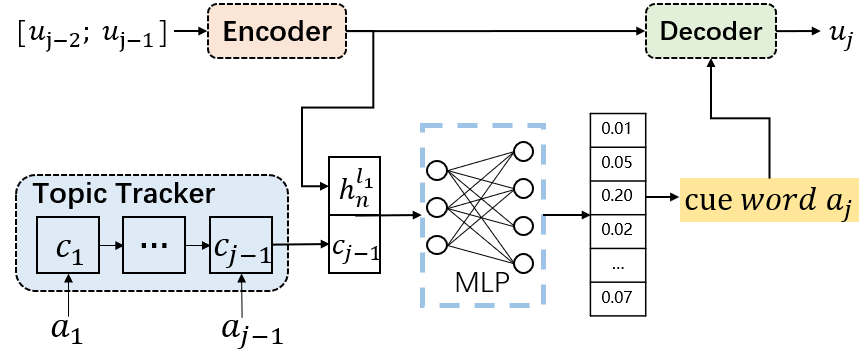}
  \caption{The end-to-end framework for cue words selection and topic augmented response generation.}
  \label{framework}
\end{figure}


\subsection{Policy Model: Cue Words Selection}

The policy model is designed for cue words selection. Given the dialogue state, we choose a cue word based on current policy. Then, we estimate the expected reward and optimize the parameters of the policy network.

Inspired by prior work~\cite{li2016deep,lewis2017deal}, we simulate two virtual conversational agents talking with each other. During simulation, we firstly use a message $Q_{i,j}$ sampled from training data to initialize the dialogue session. Then, two chatbots take turns to encode dialogue history and predict a cue word to express the main gist of the upcoming utterance. Based on the selected cue word, a response is generated until $T$ turns. Formally, a simulated conversation $\mathcal{C}$ is the combination of dialogue session and cue words sequence:
\begin{equation}
    \begin{aligned}
    \mathcal{C}&=\left(\mathcal{S'},\mathcal{F'}\right) \\
    \mathcal{S'}&=\left(Q_{i,j}, u_{i,j}, ..., u_{i,j+T-1}\right) \\
    \mathcal{F'}&=\left(\mathcal{F}_{i,j}, a_{i,j}, ..., a_{i,j+T-1}\right) \\
    Q_{i,j} &= [u_{i,j-2}, u_{i,j-1}] \\
    \mathcal{F}_{i,j} &=\left(a_{i,1}, a_{i,2}, ..., a_{i,j-1}\right) \\
    \end{aligned}
\end{equation}
where $i$ is the training instance index. Both dialogue session $\mathcal{S'}$ and cue words sequence $\mathcal{F'}$ consist of history and simulation information.

\subsubsection{State} We explore \textit{context tracker} and \textit{topic tracker} to depict dialogue state.
As for context tracker, we only focus on the previous two dialogue utterances due to the computational complexity on modeling the long-term dialogue history. To be specific, it is the encoded vector representation of the previous two utterances.

\begin{equation}
\label{context tracker}
h_n^{l_1} = \begin{cases}
\text{$\textsc{Enc}\left(Q_{i,j}\right)$} & \text{$k=j$}\\
\text{$\textsc{Enc}\left([u_{i,k-2}; u_{i,k-1}]\right)$} & \text{$k>j$}
\end{cases}
\end{equation}
where $\textsc{Enc}\left(\cdot\right)$ denotes for sequence encoding, $h_n^{l_1}$ is final hidden state of the first layer in Eq.~\ref{encoder}.

To model a natural and coherent conversation flow, we present topic tracker to represent the topic flow.
\begin{equation}
\label{cue word LSTM}
\begin{aligned}
c_{i,k-1} = \begin{cases}
\text{$\textsc{Enc}\left(\mathcal{F}_{i,j}\right)$} & \text{$k=j$}\\
\text{$\textsc{LSTM}\left(c_{i,k-2},a_{i,k-1}\right)$} & \text{$k>j$}
\end{cases}
\end{aligned}
\end{equation}
where $c_{i,k}$ is the hidden state of topic tracker. Note that we do not share the sequence encoding parameters in Eq.~\ref{context tracker} and Eq.~\ref{cue word LSTM}. Further, the dialogue state is given by: 
\begin{equation}
\label{state equation}
    s_{i,k} = [h_n^{l_1}, c_{i,k-1}]
\end{equation}
In this way, we describe the dialogue state comprehensively.



\subsubsection{Action} Given the current conversation state, an action $a_{i,k}$ is a cue word to select. Usually, the cue word is enlightening, which drives the generated response to a specific direction. Different from~\newcite{li2016deep}, we aim to optimize cue words selection so that these appropriate cue words shape the conversation flow and make the dialogue more informative and attractive. 



\subsubsection{Policy} In our reinforcement learning based conversation model, the policy is defined as the probability distribution over the action space. 
Specifically, based on the current dialogue state, we calculate the probability distribution over pre-defined cue words vocabulary. The one with the highest probability will be selected as current cue word.

\begin{equation}
\label{cue word selector}
\begin{aligned}
p\left(a_{i,k} | s_{i,k}\right) = \textit{softmax}\left(\tanh\left(W_a s_{i,k} + b_a\right)\right)
\end{aligned}
\end{equation}
where $W_a$ and $b_a$ are weight matrices and bias term. To speed up the training and avoid language divergence, we fix the encoder and decoder parameters and only optimize the policy model during reinforcement learning.

\subsubsection{Reward} The reward $r_{i,k}$ indicates the contribution of an action $a_{i,k}$ to the success of a conversation. We aim to measure a good cue word from different aspects. 

1) Effectiveness. A suitable cue word should be related to current dialogue state and reply generation, i.e., the generated reply should be semantically relevant to the predicted cue word, no matter whether the cue word explicitly appears in the reply or not. The reward is given by the log cosine similarity between them:
\begin{equation}
\begin{aligned}
r_1 = \log cos\left(a_{i,k},u_{i,k}\right)\cdot cos\left(a_{i,k},u_{i,k-1}\right) 
\end{aligned}
\end{equation}
We adopt an embedding-based metric~\cite{liu2016not} to measure the correlation between the predicted cue word and dialogue sentences (current dialogue history or generated reply). We do not compute sentence-level embeddings; instead, the cue word is greedily matched with each token in a dialogue sentence based on the cosine similarity of their word embeddings. The highest cosine score is regarded as the correlation between them.


2) Relevance. The basic requirement of a dialogue system is that the generated response should be related to dialogue context. The relevance reward is given as follows:
\begin{equation}
\begin{aligned}
r_2 = m\left(u_{i,k},\left(u_{i,j-2},u_{i,j-1},...,u_{i,k-1}\right)\right)
\end{aligned}
\end{equation}
where $m(\cdot)$ is a pre-trained multi-turn conversation matching network~\cite{wu2016sequential}. We adopt the matching score to measure the relevance between a response and dialogue context. 

To sum up, the reward for action $a_{i,k}$ is defined as: 
\begin{equation}
\begin{aligned}
r_{i,k} = \alpha r_1 + \left(1-\alpha\right) r_2
\end{aligned}
\label{reward}
\end{equation}
where $0 < \alpha < 1$ and we set $\alpha = 0.2$. To estimate the expected reward, we take the future influence into consideration:
\begin{equation}
\begin{aligned}
\mathbb{E}\left[r\left(a_{i,j}, s_{i,j}\right)\right] = \sum_{k=j}^{j+T-1}r_{i,k}\cdot\gamma^{k-j}
\end{aligned}
\label{expectedreward}
\end{equation}
where $\gamma$ is decay factor and we set $\gamma=0.9$. 
The learning process iteratively estimates and maximizes the expected future rewards.

\subsection{Model Training}
\label{training}
  To warm up for the policy model, we firstly fit our model to human-human conversational patterns with supervised learning. Then, through the simulated dialogues between two agents, we optimize the policy model with reinforcement learning.

  \begin{algorithm}[t]
      \caption{Training Process}
      \label{trainingprocess}
      \begin{algorithmic}[1]
      \State $N$: number of training instances.
      \State $T$: simulation turns.
      \State $M$: sampling times.
      \State $L_i$: number of utterances in $i$-th dialogue session.
      \State Jointly pre-train policy model and cue word augmented response generation model with supervised learning.
       \For {$i=1...N$}
          \For {$j=1...L_i$}
              \State $s_{i,j}=[c_{i,j-1}; h_n^{l_1}]$;
              \For {$m=1...M$}
                \For {$k=1...{j+T-1}$}
                    \State Sample an action $a_{i,k}$ based on $s_{i,k}$
                    \State Compute $r_{i,k}$ using Eq.~\ref{reward}
                    \State Transform to state $s_{i,k+1}$
                \EndFor
                \State Estimate $\mathbb{E}\left[r\left(a_{i,j}, s_{i,j}\right)\right]$ using Eq.~\ref{expectedreward}
                \EndFor
            \State Compute average reward $b_{i,j}$
            \State Update policy model $\mathcal{P}$ using Eq.~\ref{gradient}.
          \EndFor
       \EndFor
      \end{algorithmic}
  \end{algorithm}

  \subsubsection{Single-turn Supervised Learning} 
  \label{supervised}
  For the first stage of training, we aim to learn the cue word augmented response generation model. Given an aligned (query, cue words, response) tuple $(Q_{i,j},\mathcal{F}_{i,j}, \hat{R}_{i,j})$, 
  we sample a meaningful word (noun, verb, or adjective) from the reference response $\hat{R}_{i,j}$ as the gold cue word $\hat{a}_{i,j}$. During training, we firstly select a cue word $a_{i,j}$ with highest probability based on dialogue context $Q_{i,j}$ and topic flow $\mathcal{F}_{i,j}$. Then, we generate a reply $R_{i,j}$. Formally, the cue word augmented generation model can be formulated as:
 
  \begin{equation}
  \begin{aligned}
  \mathcal{G}\left(R_{i,j}\right) &= p\left(R_{i,j}\mid Q_{i,j}, \mathcal{F}_{i,j}, {a}_{i,j}\right) \\
  {a}_{i,j}&=\argmax_{{a}_{i,j}} p\left({a}_{i,j}|{Q}_{i,j}, \mathcal{F}_{i,j}\right)
  \end{aligned}
  \end{equation}
  

Naturally, the objective function is to minimize the cross entropy of cue words selection and responses generation:
  \begin{equation}
  \begin{aligned}
  \mathcal{D}=-\sum_{i=1}^{N}\sum_{j=1}^{L_i}\left[{a}_{i,j}\log p\left(\hat{a}_{i,j}\right) + \sum_{t=1}^{m}y_{i,j,t}\log p\left(\hat{y}_{i,j,t}\right)\right]
  \end{aligned}
  \end{equation}
  where $N$ is the number of training instances, $L_i$ denotes the number of utterances in the $i$-th dialogue session, $m$ indicates the length of reply words. $y_{i,j,t}$ is the one-hot representation of $t$-th word in reply $R_{i,j}$.

  \subsubsection{Conversation Simulation} 
  \label{simulation}
  To train the policy model, we firstly initialize it with the supervised model mentioned above. Then, we optimize it with conversation simulation. 
  
  The simulation process between two chatbots (sharing same parameters) consists of following steps: 1) An initial instance $({Q}_{i,j},\mathcal{F}_{i,j})$ from training data is fed to agent A as input. 2) Agent A samples a cue word $a_{i,j}$ based on the policy model, which computes the probability distribution over cue words vocabulary. 3) Given dialogue context and the selected cue word, agent A generates a response $u_{i,j}$. 4) Transform to new dialogue state $s_{i,j+1}$ using Eq.~\ref{state equation},
  which is fed to agent B as input. 5) Repeat from step 2) to 4) until the conversation reaches an end.  

  We define the maximum simulation turns $T$ to terminate the simulation process. During simulation, we aim to optimize the policy parameters to improve the probability of an action with a greater expected reward. We adopt policy gradient~\cite{williams1992simple} for optimization. The objective of learning is to maximize the expected future reward:

\begin{table}[t]
  \centering
    \begin{tabular}{cc} \hline
    \toprule[0.5pt]
      Number of sessions & $162,143$ \\
      Average number of turns & $6.79$ \\
      Vocabulary Size & $32,979$ \\
      Average length of utterances & $7.79$ \\ 
    \bottomrule[0.5pt]
    \end{tabular}%
  \caption{Statistics of the Weibo dataset after filtering.}
  \label{statistics}   
\end{table}

  \begin{equation}
  \mathcal{J}\left(\theta\right)=\mathop{\mathbb{E}_{data\sim \mathcal{D}, a_{\cdot,\cdot}\sim \mathcal{P}}}[r\left(a_{\cdot,\cdot},s_{\cdot,\cdot}\right)]
  \end{equation}
  
  The gradient of objective function is calculated by REINFORCE algorithm~\cite{williams1992simple}:
  \begin{equation}
  \small
  \begin{aligned}
    \nabla_{\theta}\left(\mathcal{J}\right) \approx \frac{1}{N}\sum_{i=1}^{N}
    \frac{1}{L_i}\sum_{j=1}^{L_i}\nabla_{\theta}\log
    p\left(a_{i,j}|s_{i,j}\right)\cdot \\\left(\sum_{k=j}^{j+T-1}r_{i,k}\cdot\gamma^{k-j}-b_{i,j}\right)
  \end{aligned}
  \label{gradient}
  \end{equation}

  where $b_{i,j}$, the average reward of different sampling actions in the same state $s_{i,j}$, is a bias estimator to reduct variance:
  \begin{equation}
    b_{i,j}=\frac{1}{M}\sum_{m=1}^{M}r\left(a_{i,j}^{m},s_{i,j}\right)
  \end{equation}
  where $a_{i,j}^{m}$ is the $m$-th sampling action in the same state $s_{i,j}$. Together with supervised learning, the whole training process is summarized in Algorithm~\ref{trainingprocess}.

\section{Experiments}

\begin{table*}[t]
    \centering
    \begin{tabular}{c|c|c|c|c|c|c|c|c|c|c|c}
        \toprule[1.0pt] 
        \multirow{2}{*}{Method} & \multirow{2}{*}{Turns} & \multicolumn{3}{c|}{Intra-session} & \multicolumn{3}{c|}{Inter-session} & \multirow{2}{*}{\# U.} & \multirow{2}{*}{\# B.} & \multirow{2}{*}{\# T.} & \multirow{2}{*}{\# Words} \\ \cline{3-8}
        & & Dist-1 & Dist-2 & Dist-3 & Dist-1 & Dist-2 & Dist-3 & & & &  \\ \hline 
        S2S & 2.57 & 0.52 & 0.52 & 0.41 & 0.01 & 0.05 & 0.10 & 7.83 & 9.65 & 8.63 & 2,435 \\
        S2S-Cw & 4.38 & 0.52 & 0.57 & 0.46 & 0.01 & 0.07 & 0.16 & 11.74 & 16.36 & 15.20 & 4,733  \\
        RL-S2S & 5.45 & \bf 0.58  & \bf 0.66 & 0.54 & 0.01 & 0.05 & 0.11 & 18.91 & 24.31 & 20.67 & 4,219 \\ \hline

        RLCw-E. & 5.93 & 0.50 & 0.61 & 0.53 & 0.01 & 0.07 & 0.18 & 16.78 & 24.95 & 23.86 & 4,889 \\ 
        RLCw-R. & 6.30 & 0.53 & 0.65 & \bf 0.56 & 0.01 & 0.07 & 0.19 & \bf 19.94 & \bf 29.69 & \bf 28.30 & \bf 5,726 \\
        RLCw & \bf 6.51 & 0.52 & 0.64 & 0.55 & 0.01 & \bf 0.08 & \bf 0.20 & 19.43 & 28.95 & 27.44 & 5,637  \\
        \bottomrule[1.0pt]
    \end{tabular}%
    \caption{Automatic evaluation results of our proposed model against baselines. Suffix ``-E.'' and ``-R.'' denote the RLCW model only with the reward of effectiveness or relevance, respectively.  Turns refers to the average number of simulated turns. \# U., \# B., and \# T. are the average numbers of distinct unigram, bigram, and trigram in a dialogue session. \# Words denotes the number of distinct words in all simulated conversations.}
    \label{diversity}   
\end{table*}

\label{experiment}
In this section, we compare our method with three representative baselines based on a huge publicly available conversation resource. The objectives of our experiments are to 1) evaluate the effectiveness of our proposed RLCw model, and 2) explore how selected cue words affect the dialogue process.

\subsection{Dataset}
\label{dataset}

We conduct experiments on a public multi-turn Weibo dataset\footnote{\url{http://tcci.ccf.org.cn/conference/2018/dldoc/trainingdata05.zip}
}, which consists of $5,000,000$ and $40,000$ conversation sessions in training and testing set, respectively.

The datasets are collected from Sina Weibo~\footnote{~\url{https://www.weibo.com/}}, one of the most popular social media sites in China, used by
over 30\% of Internet users~\cite{wang2018chat}, covering rich real-world topics in our daily life. To ensure higher data quality, we construct the experimental dataset in the following steps: 1) Keep the conversational sessions with more than two turns. 2）Remove repetitive training instances\footnote{For a dialogue session, we extract consecutive two utterances as a query and the following one as the reply. Any empty sentence is not allowed.}. 3) As for the instances with the same reply, we only use ten of those with the most query words.
4) We build a vocabulary of noun, verb, and adjective\footnote{We use \textit{Jieba} as our segmentation and POS tagging toolkit. \url{https://github.com/fxsjy/jieba}}. Then, we keep top 999 frequent words and a special symbol $\left\langle \textsc{EPT}\right\rangle$ as the cue words set. For each utterance, we match the longest word from the cue words set. If not, it is labeled as $\left\langle \textsc{EPT}\right\rangle$. And we only maintain 1000 training instances with the special label. 5) The special symbol $\left\langle \textsc{UNK}\right\rangle$ will replace these words whose frequency is less than 11 times in training data. Table~\ref{statistics} presents the statistic of experimental Weibo dataset after filtering. Further, we split it into 8:1:1 for training, validation, and testing.

\begin{table*}[t]
    \centering
        \begin{tabular}{c|l|c|c|c|l|c|c|c|l|c|c|c}
            \toprule[1.0pt] 
            \multirow{2}{*}{Choice \%} & \multicolumn{4}{c|}{RLCw \textit{vs} S2S} & \multicolumn{4}{c|}{RLCw \textit{vs} S2S-Cw} & \multicolumn{4}{c}{RLCw \textit{vs} RL-S2S}  \\ \cline{2-13}
            & RLCw & S2S & Tie & Kap. & RLCw & S2S-Cw & Tie & Kap. & RLCw & RL-S2S & Tie & Kap. \\ \hline 
            Fluency & \textbf{48.0}$^{**}$  & 23.5 & 28.5 & 0.43 & \textbf{38.8}$^{**}$  & 27.3 & 33.9 & 0.42 & \textbf{41.2}$^{*}$  & 32.5 & 26.3 & 0.40 \\ 
            Consistency & \textbf{48.2}$^{**}$ & 25.7 & 26.1 & 0.39 & \textbf{38.2}$^{*}$  & 30.8 & 31.0 & 0.42 & \textbf{39.0}$^{*}$  & 32.0 & 29.0 & 0.43 \\ 
            Relevance & \textbf{37.0}$^{**}$  & 27.2 & 35.8 & 0.43 & \textbf{34.7}$^{**}$  & 26.2 & 39.1 & 0.46 & \textbf{34.3}  & 30.7 & 35.0 & 0.42\\ 
            Informativeness & \textbf{61.5}$^{**}$ & 19.7 & 18.8 & 0.40 & \textbf{51.3}$^{**}$  & 26.2 & 22.5 & 0.41 &\textbf{51.6}$^{**}$ & 25.8  & 22.6 & 0.39\\ 
            Preference & \textbf{37.7}$^{**}$  & 20.2 & 42.1 & 0.46 & \textbf{35.8}$^{**}$ & 24.3 & 39.9 & 0.44 & \textbf{34.7}$^{*}$  & 28.2 & 37.1 & 0.42 \\ 
            \bottomrule[1.0pt]
        \end{tabular}
    \caption{Human evaluation results on five aspects: fluency, consistency, relevance, informativeness, and overall user preference. We conducted significance test (t-test); ** and * indicate $p <$ 0.01 and 0.05, respectively. Kap. denotes Kappa coefficient, which shows moderate agreement among evaluators.}
    \label{human-eval}   
\end{table*}

\subsection{Baselines}
In this paper, we conduct experiments to compare our proposed method against three representative baselines.

\textbf{S2S} We implemented sequence to sequence generation model~\cite{venugopalan2015sequence}, which is treated as a preliminary baseline.

\textbf{S2S-Cw} In cue word augmented response generation method, we jointly model cue words selection and responses generation. Without future reward optimization, the supervised training process is based on the pre-trained S2S model.

\textbf{RL-S2S} Based on the pre-trained S2S model, RL-S2S~\cite{li2016deep} further optimize it with reinforcement learning. To do this, the model explored the action space (to generate a dialogue utterance) and learned the policy, which takes the form of an LSTM encoder-decoder. 

Different from RL-S2S baseline, our proposed RLCw model focuses on adaptively cue words selection so as to lead the conversation direction naturally.

\subsection{Experiment Setup}
As all our baselines and the proposed method are RNNs-based generative models, we conduct the same set of hyper-parameter optimized for them. In our experiments, the batch size was 64. We limited the maximum length of a source input to 44 words (22 for output) and directly discarded the part that exceeds the length constraint. During training and testing, the numbers of simulated conversational turns were set to 3, 10 respectively.

The dimension of word embedding was set to 600. To speed up training, we pre-trained Word2Vec~\cite{mikolov2013distributed} using \textit{gensim} library\footnote{\url{https://radimrehurek.com/gensim/models/word2vec.html}}. Both LSTMs have 1000 hidden cells for each layer. We applied Adam~\cite{kingma2014adam} to optimize the objective function and the learning rate was initialized as 0.0001. These values were mostly chosen empirically.

\begin{table*}[t]  
    \centering
    \begin{tabular}{l|p{7cm}|p{7cm}}
        \toprule[1.0pt] 
        Source Input &  等你出道 & I'm waiting for your official debut. \\ \hline
        Cue words & 美 $\to$ 机智 $\to$安逸 $\to$ 发型 $\to$ 兄弟 $\to$ 撩& goddess $\to$ witty $\to$ comfort $\to$ hair style $\to$ brother $\to$ attract \\ \hline
        \multirow{6}{*}{Dialogue} & A: 你的\underline{美女}是谁 & A: Who is your \underline{goddess}?\\
        & B: \underline{机智}的你  & B: It's you! My \underline{cute} girl.\\
        & A: 你最棒，我是最漂亮的  & A: You are the best, and I am the most beautiful.\\
        & B: 你的头像是我吗  & B:  Do you use my photo as portrait? \\ 
        & A: \underline{兄弟}，我是你的小粉丝  & A: \underline{Brother}, I'm your fan.\\
        & B: 你是不是被我\underline{撩}走了  & B:  Are you \underline{attracted} by me?\\
        \hline \hline
        \multirow{2}[0]{*}{Source Input}  &  \multirow{2}[0]{*}{我能说我高中三年都是球头吗。这是学校规定} & I have been with buzz cut for three years in high school. This is one  of the school regulations.\\ \hline
        Cue words & 美女 $\to$ 做 $\to$ 找 $\to$ 重庆& beauty $\to$ work $\to$ visit $\to$ Chongqing\\ \hline
        \multirow{4}{*}{Dialogue} & A: 哦哦，这样啊，\underline{美女}，你今天回去了吗 & A: Oh, I see. \underline{Beauty}, did you go back today?\\
        & B: 明天回了，我在\underline{做}毕业设计 & B: I will go back tomorrow. I'm \underline{working} on my graduation thesis. \\
        & A: 我明天可以去 \underline{找}你嘛，你去哪了 & A: May I \underline{visit} you tomorrow? By the way, where have you been?\\
        & B: 我去南京& B: Nanjing.\\ 
        \bottomrule[1.0pt]
    \end{tabular}%
    \caption{Case studies of the generated cue words and dialogues.}
    \label{casestudy}
\end{table*}

\subsection{Evaluation Metrics}
We evaluate different methods with both automatic metrics and human judgments.

\subsubsection{Automatic metrics.} Inspired by the simulation strategy for training the policy model, we also use the simulation during testing. The system needs high-quality initial input to start the conversation, since it is unclear how to proceed the dialogue with the beginning of ``me too''. Therefore, we manually build a set of dull sentences to further filter test data with a meaningless query. Totally, there are 17,832 input messages for testing.

The first automatic metric we use is the average turns of the simulated dialogue. We define termination conditions as follows: 1) A dull sentence is generated. 2) There are more than 80\% overlap of words between two consecutive utterances from the same or different agents. 3) Simulation turns reach the maximum limit during testing. This metric is employed to measure the conversational engagement of different methods.

The degree of diversity is another important measurement for conversation systems. We compute the ratio of distinct unigram, bigram, and trigram in the generated utterances, which are denoted as Dist-1, Dist-2, and Dist-3, respectively. To show the fine-grained difference, we report the diversity in intra- and inter-session level.

\subsubsection{Subjective metrics.} 
We also conduct pairwise human evaluation to assess subjective quality of generated multi-turn conversations. Given two simulated dialogues, we compare them from five aspects: 
\textit{fluency} (the generated sentences are fluent without grammatical errors), \textit{consistency} (whether the conversation is logically consistent and coherent), \textit{relevance} (whether the responses are semantically relevant to dialogue context),
\textit{informativeness} (whether the dialogue is informative and meaningful)
and \textit{overall user preference} (how do users like the dialogues).

\subsection{Main results}
\subsubsection{Automatic evaluation.} The automatic results of our model against all baselines are listed in Table~\ref{diversity}. As we see, our proposed RLCw model significantly outperforms baselines in simulation turns, which demonstrates more active engagement of our method. Besides, our proposed RLCw model generates more diverse outputs; it obtains the highest ratio of distinct trigram in both intra- and inter-session level. Besides, our RLCw model is slightly inferior to baselines in Dist-1 metric, mainly because of more simulation turns. 

As for the baselines, the performance of S2S model is not as good as others. S2S-Cw is slightly inferior to RL-S2S model. However, it generates more distinct words in all simulated conversations comparing with RL-S2S model, as the augmented cue words provide it a broader space for learning. Our proposed RLCw model absorbs its advantage actively and flows a longer and more diverse conversation.

To verify the effectiveness of proposed rewards, we also conducted an ablation study. From Table~\ref{diversity} 
we see that, both RLCw-E. (only use effectiveness reward) and RLCw-R. (only use relevance reward) outperform baselines. Comparing with RLCw-E. method, RLCw-R. tends to flow longer and more diverse conversations, which demonstrates the importance of quality constraint in responses generation. Together with these two rewards, RLCw obtains comparable diversity performance and the longest simulation turns, which reflecting the highest user engagement.

\subsubsection{Human evaluation.} we randomly sample 150 messages from test data to conduct a pairwise comparison, i.e., given an input message, we group two simulated dialogues together\footnote{For fairness, two simulated conversations are pooled and randomly permuted.} and ask the evaluators to choose which is better. We invited four native speakers to offer a judgment. The results of human evaluation against all baseline methods are listed in Table~\ref{casestudy}. Like the automatic evaluation results, RLCw consistently outperforms other baselines, which demonstrates the effectiveness of our proposed method. Especially, our proposed method shows prominent improvement in term of informativeness.

\subsection{Analysis}
We have elaborated the overall performance of all methods in the last subsection. Next, we will look closer into how cue words affect the dialogue process. 

\subsubsection{Cue words analysis.} First, we measure the quality and impact of the generated cue words sequence. We try to estimate the quality based on the average cosine similarity of each word pair in the cue words sequence. The result is 0.096 (the correlation of extracted cue words from reply sentences in training data is 0.137), which shows the semantic compactness of them. Again, we use embedding-based metric~\cite{liu2016not} to estimate the correlations between a cue word and the corresponding generated response. The correlation score is 0.81. As we found that there are about 41\% cue words appearing in the simulated dialogues, the selected cue words have a great impact on response generation.

\subsubsection{Case study.} We further present two representative examples of our generated dialogues in Table~\ref{casestudy}. In the first example, our system dynamically plans a fluent dialogue flow. Based on selected cue words, our RLCw model generates coherent and interesting dialogues. In the second example, our system firstly responds to the given input message, and then shift the topic to ``beauty'' with cue words augmentation, which further affects the direction of the follow-up dialogues.


\section{Conclusion}
\label{conclusion}
We study open domain dialogue generation with cue words augmentation which leads the direction of conversations. Specifically, we present the multi-turn cue-words driven conversation system with reinforcement learning (RLCw), which jointly models the cue word prediction and response generation in an end-to-end framework. To select higher quality cue words, we design a new reward to measure the effectiveness and relevance of cue words. 
We conduct experiments on a publicly available dataset to evaluate our model on dialogue duration, diversity as well as human judgements, showing that the proposed method consistently outperforms a set of competitive baselines.

\bibliography{aaai2019}
\bibliographystyle{aaai}

\end{CJK*}

\end{document}